# Skin cancer reorganization and classification with deep neural network

Hao Chang[1]

1. *Department of Genetics, Yale University School of Medicine*
2. *Email: changhao86@gmail.com*

**Abstract**

As one kind of skin cancer, melanoma is very dangerous. Dermoscopy based early detection and recognition strategy is critical for melanoma therapy. However, well-trained dermatologists dominant the diagnostic accuracy. In order to solve this problem, many effort focus on developing automatic image analysis systems. Here we report a novel strategy based on deep learning technique, and achieve very high skin lesion segmentation and melanoma diagnosis accuracy: 1) we build a segmentation neural network (skin_segnn), which achieved very high lesion boundary detection accuracy; 2) We build another very deep neural network based on Google inception v3 network (skin_recnn) and its well-trained weight. The novel designed transfer learning based deep neural network skin_inceptions_v3_nn helps to achieve a high prediction accuracy.

**Introduction**

Melanoma is the most dangerous type of skin cancer. It kills more than 10,000 people each year in United States[1]. Early detection, reorganization and treatment is critical for melanoma therapy. It can help save more than 95% people. Dermoscopy is one of the most important techniques to exam skin lesions and can capture high resolution images of the skin escaping the interruption of surface reflections. Specially well-trained clinicians use this high resolution imaging to evaluate the possibility of melanoma at the very beginning and can obtain a diagnostic accuracy as high as ~80%[2]. However, there is not enough experienced dermatologists all over the world. In order to solve this problem, there has been effort to create machine-driven image analysis software to classify different skin associated diseases using dermoscopy images in the academic research community. Previous computer-aided classification strategies are less successful for two major reasons: 1) the database is not as good as enough. Previous work is based on quite few amount of dermoscopy skin lesion images, so programs cannot learn and extract useful features. 2) There was also limited computing ability in previous time, so people had little ideas to treat with a huge amount of images.

Artificial intelligent and deep learning technique, powered by advanced computation ability and large datasets people have collected and published as open sources, have been dominated in many areas and been proved to exceed human performance in strategic games like Go[3], image recognition like ImageNet competition[4], language translation and speech recognition. A very important paper published recently in nature also validates that deep constitutional neural network exhibits very high melanoma classification ability and has been shown to exceed common well trained dermatologists (72.1% vs 66.0%)[5]. In this paper, the author utilized a GoogleNet Inception v3 CNN architecture with well trained weight on 1.28 million ImageNet datas. The final layer is removed and fine tune to the author categorized dataset containing more than 13,000 images which was collected from a combination of open-access dermatology repositories, the ISIC Dermoscopic Archive, the Edinburgh Dermofit Librar. The significance of this paper is that it is the first time scientists proved the diagnostic ability of deep learning in skin cancer screen. This is also the major concept of ISIC Dermoscopic Archive dataset and many other open source medical imaging datasets. Those datasets help us to improve the whole research community and to develop more powerful artificial intelligent based methodologies. In the future, we believe that artificial intelligent and deep learning would obtain more powerful results and finally benefit the whole society and human health.

In this ISIC skin cancer competition, I am trying to develop a novel deep learning based solution to solve melanoma classification problem. In details, I am trying to solve two major problems: 1) the skin lesion recognition problem (ISIC Challenge Part 1). Skin lesion recognition is very important, because it would be helpful for cameras and computers to automatically detect and zoom-in to the lesion regions. I have tried many strategies and finally chose the U-Net[6] based neural network and have achieved a good detection and recognition accuracy. 2) the melanoma judgment (ISIC Challenge Part 3). I utilize the transfer learning strategy based on Google well trained Inceptions_v3 neural network. This novel designed network also help to achieve a very high melanoma prediction accuracy.

**Results**

**Datasets.** My training data is downloaded from the ISIC Challenge website. Total 2000 dermoscopic images includes 374 melanoma images, 1372 nevus images and 254 seborrheic keratosis images (**Figure 1A**). In Part 1 lesion segmentation competition, the hand drawn segmentation binary images are also provided. The validation data includes 150 unlabeled images and the final testing data contains 600 unlabeled images. The ISIC Challenge also provided the superpixel images and some basic information of patient including age and sex. In my training and testing neural network, I gave up the utility of patient information. It is proved that melanoma is age and sex dependent by analyzing the ISIC datasets and cancer society collected (**Figure 1B-C**). However, in my strategy, this two parameters is eliminated.

**Image Preprocessing.** All the images are resized into 150X150 for both skin_segnn and skin_recnn. Three channels of the dermoscopic images (Red, Green, Blue) are utilized in both neural network two. I also normalized the input images by subtracting the mean pixel intensity and dividing the standard deviation. This treatment makes to image mean value become zero and standard deviation become 1. The input data of skin_segnn is the 2000 treated images. But the input data of skin_recnn is not only the 2000 treated images but also added the cropped skin lesion images from the datasets, because high definition images are very important for dermatologists to clarity. For validation and testing, the cropped images is obtained from the segmentation results of the skin_segnn.

**Segmentaion Net.** The skin lesion segmentation neural network (skin_segnn) is similar like U-Net architecture. It is powered by deep learning package Tensorflow, Theano and Keras. The input image is the original images and the output is the segmentation masks. The structure of skin_segnn includes a series of convolution and pooling layers. Skip connections are used as what U-Net designed. This would be helpful to increase the stability of the network and outputs (**Figure 2A**).

**Recognition Net.** The melanoma recognition neural network (skin_recnn) is built based on the backbend of Google Inceptions V3 network (**Figure 2B**). The output layer of Inceptions V3 is replaced by and full connection layers (1024 nodes). The starting weight of this network is the well-trained weight from ImageNet datasets. There are two Inceptions V3 neural networks in my design structure. One is in charge of full size dermoscopic images, the other one is in charge of cropped dermoscopic images from skin_segnn, which only maintained the major lesion region. Then both 1024 nodes are merged together and an output layer with 3 nodes is linked below. This 3 nodes represent the 3 classes including melanoma, nevus and seborrheic keratosis. We first train the network by freezing all Inception V3 layers. Then the final two blocks of Inception V3 layers is trainable but with very low decreasing rate.

**Figure 1. Three major classes of skin disease**

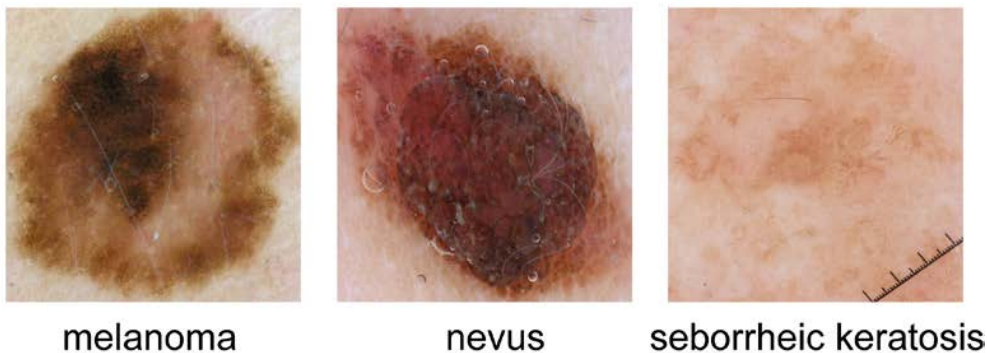

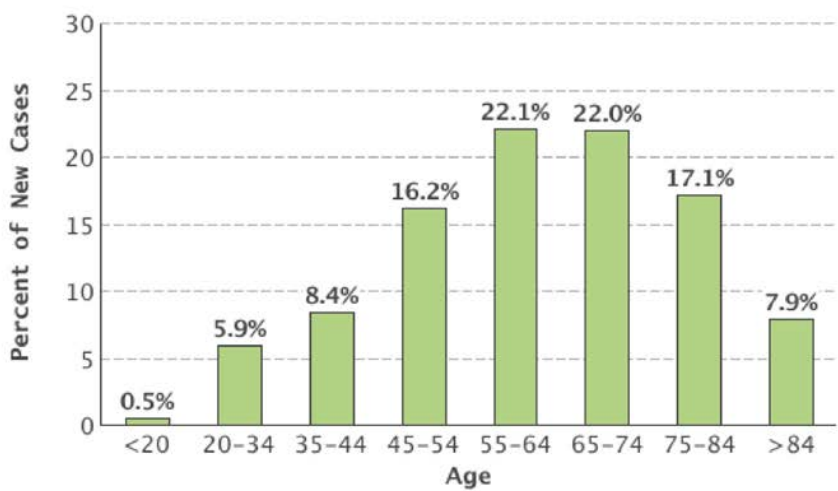

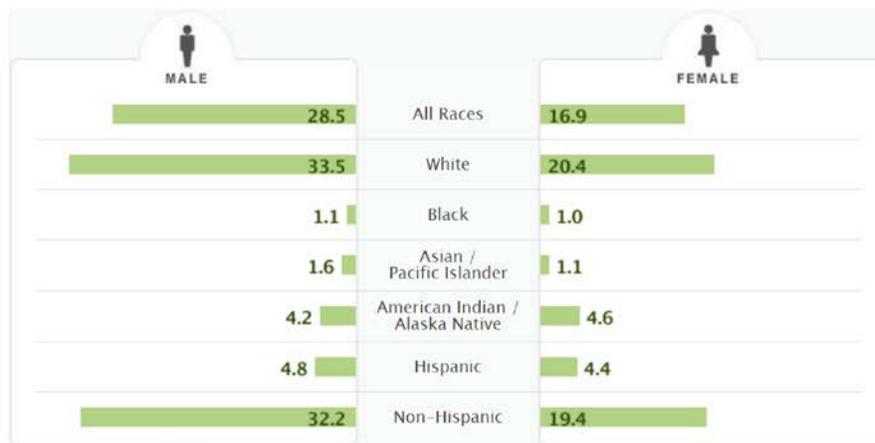

**Figure 2. The structures of skin_segnet and skin_recnet**

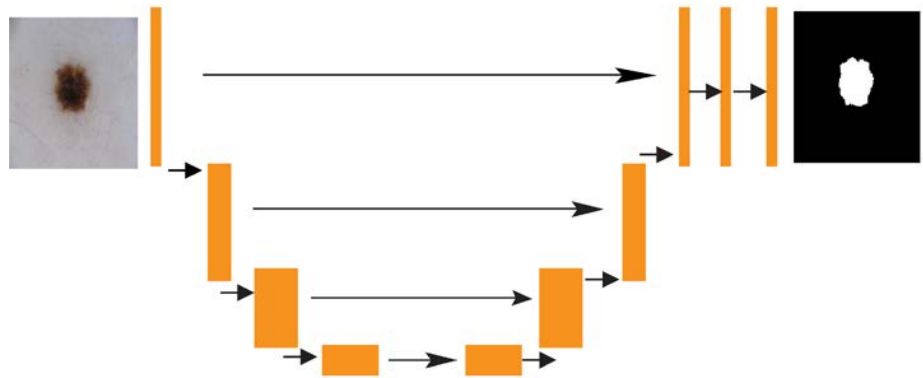

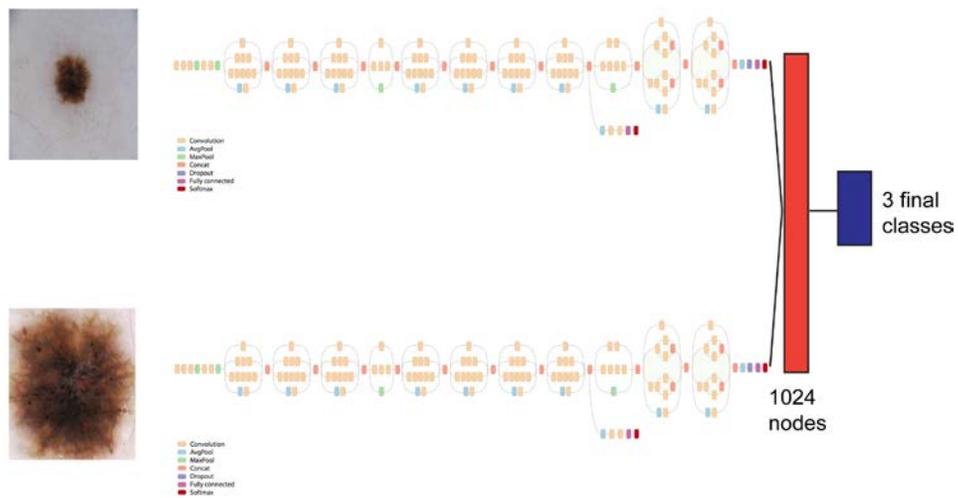

**Reference**

1. Melanoma research gathers momentum. *Lancet* **385**, 2323, doi:10.1016/S0140-6736(15)61087-X (2015).
2. Ali, A. R. A. & Deserno, T. M. A Systematic Review of Automated Melanoma Detection in Dermatoscopic Images and its Ground Truth Data. *Proc Spie* **8318**, doi:Artn 83181i 10.1117/12.912389 (2012).
3. Silver, D. *et al.* Mastering the game of Go with deep neural networks and tree search. *Nature* **529**, 484-+, doi:10.1038/nature16961 (2016).
4. Russakovsky, O. *et al.* ImageNet Large Scale Visual Recognition Challenge. *Int J Comput Vision* **115**, 211-252, doi:10.1007/s11263-015-0816-y (2015).
5. Esteva, A. *et al.* Dermatologist-level classification of skin cancer with deep neural networks. *Nature* **542**, 115-118, doi:10.1038/nature21056 (2017).
6. Ronneberger, O., Fischer, P. & Brox, T. U-Net: Convolutional Networks for Biomedical Image Segmentation. *Lect Notes Comput Sc* **9351**, 234-241, doi:10.1007/978-3-319-24574-4_28 (2015).